\documentclass{ecai}
\usepackage{latexsym}
\usepackage{times}
\usepackage{soul}
\usepackage[utf8]{inputenc}
\usepackage[small]{caption}
\usepackage{booktabs}
\usepackage[switch]{lineno}
\usepackage{amsmath}
\usepackage{amssymb}
\usepackage{amsthm}
\usepackage{mathrsfs}
\usepackage{amsfonts}
\usepackage{float}
\usepackage{multirow}
\usepackage{booktabs}
\usepackage{amsthm}
\usepackage{xspace}
\usepackage{subfig}
\usepackage{booktabs}
\usepackage{color}
\usepackage{xcolor}
\usepackage{tabularx}

\usepackage{graphicx}
\usepackage{amsmath}
\usepackage{amssymb}
\usepackage{amsthm}
\usepackage{mathrsfs}
\usepackage{algorithm}
\usepackage{algpseudocode}
\usepackage{amsfonts}
\usepackage{float}
\floatname{algorithm}{Algorithm}

\newcommand{\sysname}{\emph{SLADV}\xspace}

\begin{document}


\begin{frontmatter}

\title{On the Robustness of Split Learning against Adversarial Attacks}

\author[A]{\fnms{Mingyuan}~\snm{Fan}}

\author[A]{\fnms{Cen}~\snm{Chen}\thanks{Corresponding Author. Email: cenchen@dase.ecnu.edu.cn.}}

\author[B]{\fnms{Chengyu}~\snm{Wang}}

\author[B]{\fnms{Wenmeng}~\snm{Zhou}}

\author[B]{\fnms{Jun}~\snm{Huang}}


\address[A]{East China Normal University, China}
\address[B]{Alibaba Group, China}

\begin{abstract}
    Split learning enables collaborative deep learning model training while preserving data privacy and model security by avoiding direct sharing of raw data and model details (i.e., server and clients only hold partial sub-networks and exchange intermediate computations). 
    However, existing research has mainly focused on examining its reliability for privacy protection, with little investigation into model security. 
    Specifically, by exploring full models, attackers can launch adversarial attacks, and split learning can mitigate this severe threat by only disclosing part of models to untrusted servers.
    This paper aims to evaluate the robustness of split learning against adversarial attacks, particularly in the \textit{most challenging setting where untrusted servers only have access to the intermediate layers of the model}.
    Existing adversarial attacks mostly focus on the centralized setting instead of the collaborative setting, thus, to better evaluate the robustness of split learning, we develop a tailored attack called \sysname, which comprises two stages: 
    1) shadow model training that addresses the issue of lacking part of the model and 2) local adversarial attack that produces adversarial examples to evaluate.
    The first stage only requires \textit{a few unlabeled non-IID data}, and, in the second stage, \sysname perturbs the intermediate output of natural samples to craft the adversarial ones.
    The overall cost of the proposed attack process is relatively low, yet the empirical attack effectiveness is significantly high, demonstrating the surprising vulnerability of split learning to adversarial attacks.
\end{abstract}

\end{frontmatter}

\section{Introduction}
\label{intro}

Split Learning (SL) emerges as a promising distributed learning paradigm for addressing the privacy issue in conventional centralization training \cite{split_learning_1}. 
This learning paradigm enables collaborative model training by splitting the whole network into sub-networks that are computed by different participants (i.e., clients and a server).
It alleviates the privacy issue by keeping the sensitive raw data locally at the client side and only exchanging intermediate computations with the server.
In the training process, the server bears most of the involved computational costs, making split learning lightweight and scalable for clients with limited computing resources.

\begin{figure}[t]
    \centering
    \includegraphics[width=1.\linewidth]{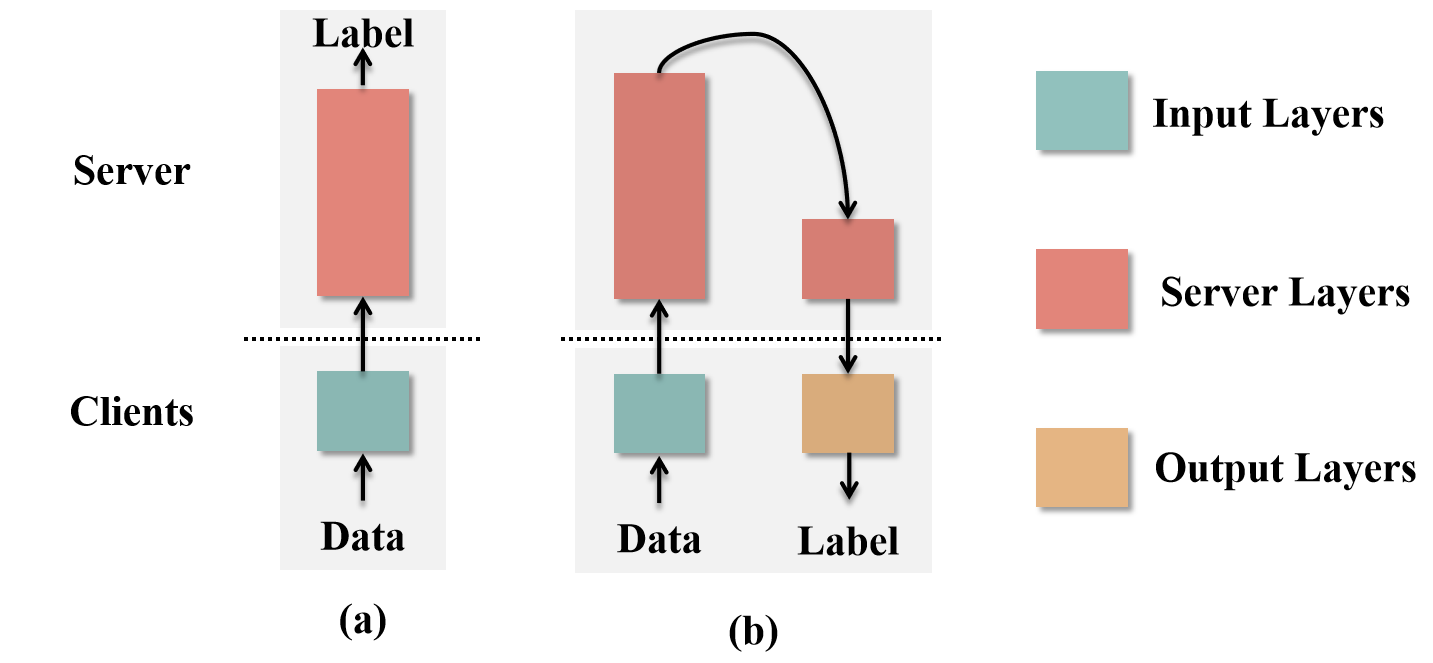}
    \caption{Sketch map of split learning: (a) vanilla split learning and (b) split learning with the U-shaped configuration.
    }
    \label{scenario_comp}
\end{figure}

Figure \ref{scenario_comp}(a) presents the \textit{vanilla split learning}~\cite{u_shaped_learning_1}, where a full model is split into two sub-networks: input layers and server layers.
Clients train the input layers up to a specific cut layer, and the resulting outputs are sent to the server. The server then completes the rest of the forward process with the server layers without accessing clients' private raw data.
The back-propagation process is also performed in a similar fashion.
However, vanilla split learning requires clients to expose labels to the server, which may compromise label privacy.
To mitigate this concern, an improved version of vanilla SL is presented in Figure \ref{scenario_comp}(b), i.e., \textit{split learning with the U-shaped configuration}~\cite{u_shaped_learning_1}, which further divides and transfers a few layers at the end of the full model (i.e., output layers) back to clients, eliminating the need for label sharing.

One of the most heated research topics of split learning is its trustworthiness and a sizable body of work focuses on examining whether split learning can truly protect privacy \cite{split_attack_1,split_attack_2}.
However, there is a lack of research on discussing model security in split learning, which also is an important aspect of trustworthiness.
In detail, attackers can manipulate models by employing
\textit{adversarial attacks}~\cite{adv_survey_2}, which generates adversarial examples by imposing human-imperceptible adversarial noises into natural samples.
Fortunately, launching adversarial attacks typically requires full access to target models, making them less dangerous in the conventional centralized learning paradigm.
In centralized learning, the trainer centralizes data across different places into one data center.
Untrusted individuals rarely are involved in the training process, thereby the models are not likely to be disclosed.
In contrast, for split learning, the server may be untrusted while part of the model is delivered to the server \cite{split_attack_1,split_attack_2}.
As a result, a key question naturally arises: can split learning maintain model security?
To address this, to our best knowledge, this paper conducts the first systematic empirical study on the vulnerability of split learning against adversarial attacks.

In this paper, we focus on examining the vulnerability of split learning with the U-shaped configuration for three reasons.
First, this type of split learning is more favorable since it allows protection for both data and labels.
Second, launching attacks in this setting is more challenging because the available resources for the server are greatly limited compared to vanilla split learning.
Third, the attacks developed for this type of split learning are more generic and can be seamlessly applied to vanilla split learning.
Thus, for simplicity, in the remainder of this paper, we use split learning to denote split learning with the U-shaped configuration (Figure \ref{scenario_comp}(b)).

Existing adversarial attacks are mainly developed in centralized settings, which cannot fully exploit the characteristics of split learning \cite{adv_survey}.
To address this issue, we first present a practical attack scenario that aligns well with split learning and then develop a novel yet effective adversarial attack called \sysname.
Adversarial attacks typically adopt input gradients as maliciously-designed noises \cite{adv_survey_3}. 
Common adversarial attacks \cite{adv_survey} assume either full access to the target model (i.e., white-box adversarial attacks) or pre-training a similar proxy model as a surrogate of the target model (i.e., transfer-based adversarial attacks).
Our proposed \sysname follows the latter type of attack, as it is more practical than the former type.

Our proposed \sysname consists of two stages: shadow model training and local adversarial attack.
In shadow model training, by leveraging the characteristics of split learning, \sysname only needs to train shadow input layers instead of a complete model.
The training of shadow input layers requires only \textit{a few hundred to a few thousand samples}, which do not necessarily have to be identical to the distribution of the target model's training data.
In contrast, the training of proxy models in common transfer-based attacks needs to collect a large amount of data that is identically distributed with the training set of the target model.
Additionally, the training in \sysname does not necessitate labeled data, eliminating the need for labeling data in common transfer-based attacks, which can be labor-intensive.
In local adversarial attack stage, the concatenation of shadow input layers and server layers serves as the proxy model to produce adversarial examples.
Extensive experiments show that \sysname achieves competitive attack performance at a significantly lower cost compared to common transfer-based attacks.
In all, we believe this study sheds light on the vulnerability of split learning to adversarial attacks, alerts clients to potential model security issues, and may inspire further defenses in this area.
Our contributions are four-fold:
\begin{itemize}
    \item To our best knowledge, we are the first to identify the potential threat in split learning, i.e., the vulnerability of split learning against adversarial attacks, and, have conducted a seminal investigation into this threat.
    \item By delving into the bottom of split learning, we have designed a highly practical attack scenario with limited attack resources.
    Within this scenario,  we developed a novel yet effective attack method named \sysname that fully exploits the characteristics of split learning.
    \item We have performed an in-depth theoretical analysis of the proposed \sysname attack and demonstrated its theoretical effectiveness under mild conditions.
    \item Extensive experiments have been conducted to examine the robustness of split learning against adversarial attacks, revealing the severe vulnerability of split learning and the effectiveness of our proposed \sysname.
\end{itemize}

\section{Related Work}
\label{preliminary}


\subsection{Model Security and Adversarial Attacks}

Although DNNs achieve impressive performance over many complicated tasks, the internal operating mechanism of DNNs is still opaque to humans and this feature can be maliciously explored to launch adversarial attacks against DNNs \cite{ai_security}.
Adversarial attacks impose little noises along model vulnerable directions into natural samples to craft adversarial examples. A lot of studies empirically show the remarkable vulnerability of DNNs against adversarial examples \cite{adv_survey,momen_attack}.
Generally, if DNNs are allowed to be fully accessible, known as the white-box scenario, the input gradient directions are considered as the model vulnerable direction \cite{FGSM,BIM}, i.e., taking gradients of loss function w.r.t. natural samples as adversarial noises.
In the conventional training paradigm, it is not likely for trained models to be disclosed to untrusted ones.
Therefore, launching attacks in the black-box scenario is more appealing for real-world applications.
Black-box attacks largely depend on the transferability of adversarial examples, i.e., adversarial examples locally crafted on a proxy model sometimes can also fool another unknown model.
Furthermore, transfer-based black-box attacks particularly follow the following two stages \cite{adv_survey,frequency_attack}: 1) collecting a sufficient amount of data together with their labels and train a proxy model from scratch, 2) locally crafting adversarial examples.
For the first stage, obtaining a similar training dataset of the target model is difficult in the real world.
Although \cite{dast2} proposed data-free proxy model training methods, the methods are limited due to the prohibitive cost of extensively querying the target model.
Our work proposes an effective attack method with limited attack resources. 

\subsection{Data Privacy and Split Learning}
Data sharing is one of the primary challenges in collaboratively training DNNs \cite{split_learning_4}. Split learning allows the training of an effective DNN without sharing any raw data \cite{split_learning_1,split_learning_2}.
In split learning, a full model is split into multiple parts, each of which is trained by a different participant.
Split learning caters to practical needs for many scenarios such as IoT settings \cite{split_learning_1} and thus is ubiquitous.
Split learning, however, typically involves an untrusted server \cite{split_attack_1,split_attack_2}.
As shown in Figure \ref{scenario_comp}(b), the full model $F_{\theta}(\cdot)$ is split into three parts, i.e., input layers $F_{\theta_1}(\cdot)$, server layers $F_{\theta_2}(\cdot)$, and output layers $F_{\theta_3}(\cdot)$.
Wherein, input layers and output layers remain in clients, and server layers are located in the server.
In each iteration, clients send the output of input layers 
into server layers. Next, the output of server layers is passed to output layers.
Clients locally compute loss and generate the gradients from output layers that propagate back to server layers and input layers.
Several recent studies \cite{split_attack_1,split_attack_2} show that the server is capable of stealing both data and labels.
Yet, to our knowledge, no related work focuses on the model security in split learning, which is our research goal to fill the gap.
\section{Threat Model}
\label{scenario}
Before developing our attack, we present the threat model used in this paper, including attacker's goal, attacker's knowledge, and attacker's capability, which defines the available resource and constraints for attacks to be performed.

\subsection{Attack Model}
\label{scenario_desc}

In split learning, multiple clients collaboratively train a model with the assistance of a server.
Following previous works \cite{split_attack_1}, we assume \textit{the server to be a malicious attacker who may intentionally exploit the training process to pursue self-interest.}
To be specific, in split learning, after training process finishes, the server delivers server layers to clients and clients concatenate input layers, server layers, and output layers into a complete model \cite{health_app_adv}.
The complete model is then deployed into real-world applications such as medical diagnosis systems \cite{health_app_adv}.
If the attacker can launch adversarial attacks to manipulate medical diagnosis systems, the attacker can fraudulently obtain government subsidies, insurance compensation, etc.

As white-box attacks necessitate full access to the target model and the server lacks access to input layers and output layers, the server can only resort to transfer-based adversarial attacks against the target model.
Our aim is to investigate the feasibility of a successful transfer-based adversarial attack in such scenario even for a highly constrained attacker.


\subsection{Attacker's Goal}
\label{attacker_goal}
The overall goal of the attacker is to launch transfer-based adversarial attacks against the target model.
Specifically, the attacker aims to deceive the target model into making false or less confident predictions for given samples, in order to achieve personal gains.


\subsection{Attacker's Knowledge}
\label{attacker_knowledge}
For the sake of the practicality of our attack, the attacker is considered to possess fairly restricted knowledge.
First, the attacker only possesses knowledge of the output generated by the input layers, as its role in split learning is solely that of a transmitter. This entails that the server cannot obtain any information pertaining to the input or output layers, including their architecture and parameters.
Second, in our SL setting, the attacker possesses only a vague understanding of the current task, rather than either complete or no knowledge. For instance, if the current task involves distinguishing between images of cats and dogs, the server may be able to deduce that it is an image-based task by analyzing the architecture of the server layers. However, the server is unlikely to be able to determine specific details about the task, such as the precise label space (e.g., cats and dogs), as the raw labeled data is not accessible to the attacker.

Note that, in common transfer-based adversarial attacks, attackers are typically assumed to possess prior knowledge of the label space of the target model in advance, enabling them to collect corresponding data associated with the label space to train a proxy model with high similarity.
Conversely, our attacker only possesses knowledge of the type of current task, i.e., an image-based task, but lacks access to the label space.
This constraint makes it unclear which images should be collected, significantly increasing the difficulty of launching attacks. Nevertheless, the restricted attacker's knowledge serves a highly practical purpose in our split learning context.

\subsection{Attacker's Capability}
\label{attacker_cap}

The attacker's capabilities in this stage adhere to the principle of avoiding behaviors that violate split learning's training rules, as such behaviors would be easily detected by clients.
As a result, we permit the following attacker's capabilities: 
1) arbitrary modifications of the input and output of server layers, which is undetectable by clients, 
2) data collection for training a proxy model with potentially dissimilar data distribution to the target model's training data\footnote{This aligns with real-world scenarios, where the attacker is assumed to have fairly limited knowledge.},
and 3) utilization of split learning's characteristics, such as input layer outputs, to train a proxy model during the training process without alerting clients. 
Notably, in existing transfer-based attacks, the proxy model's training is often delayed until the target model's training is complete. 
In contrast, in our case, the proxy model is trained together with the target model to exploit the characteristics of split learning.


\section{Approach}
\label{approach}

\begin{algorithm}[t]
  \caption{Shadow Model Training}
  \label{shadow_model_training_algorithm}
  \begin{algorithmic}[1]
    \Require $F_{\theta_1}$: input layers;
             $F_{\theta_2}$: server layers;
             $F_{\theta_3}$: output layers;
             $F_{\theta_1'}$: shadow input layers;
      $D_{1}$: dataset owned by clients;
      $D_{2}$: dataset owned by the attacker (server);
      $T$: the total number of iterations;
      $\alpha$: the hyperparameters to adjust the similarity magnitude;
      $L(\cdot,\cdot)$: a loss function.
      
    \Ensure $F_{\theta_1}$, $F_{\theta_2}$, $F_{\theta_3}$, $F_{\theta_1'}$: the trained models.

    \For{each iteration $i = 0$ to $T$}
    
        \State Clients sample $x,y$ from $D_{1}$, compute $o_1=F_{\theta_1}(x)$, and send $o_1$ to the server.

        \State The server computes $o_2=F_{\theta_2}(o_1)$ and delivers $o_2$ to clients.

        \State The server sample $x'$ from $D_2$ and computes similarity loss $L_{sim}=||F_{\theta_1'}(x')-o_1||_2$.

        \State Clients compute loss $L(F_{\theta_3}(o_2),y)$ and implement backprogation algorithm to send gradient $\frac {\partial L(F_{\theta_3}(o_2),y)}{\partial o_2}$ to the server.

        \State The server computes $g_1=\frac{\partial L(F_{\theta_3}(o_2),y)}{\partial o_2} \frac{\partial o_2}{\partial o_1}$ and $g_2=\frac{\partial L_{sim}}{\partial o_1}$.

        \State The server fuses gradients $g=g_1+\alpha g_2$ and returns $g$ to the clients.

        \State Clients and the server update model parameters based on gradients.
    
    \EndFor
  \end{algorithmic}
\end{algorithm}

\begin{figure}[t]
    \centering
    \includegraphics[width=1.\linewidth]{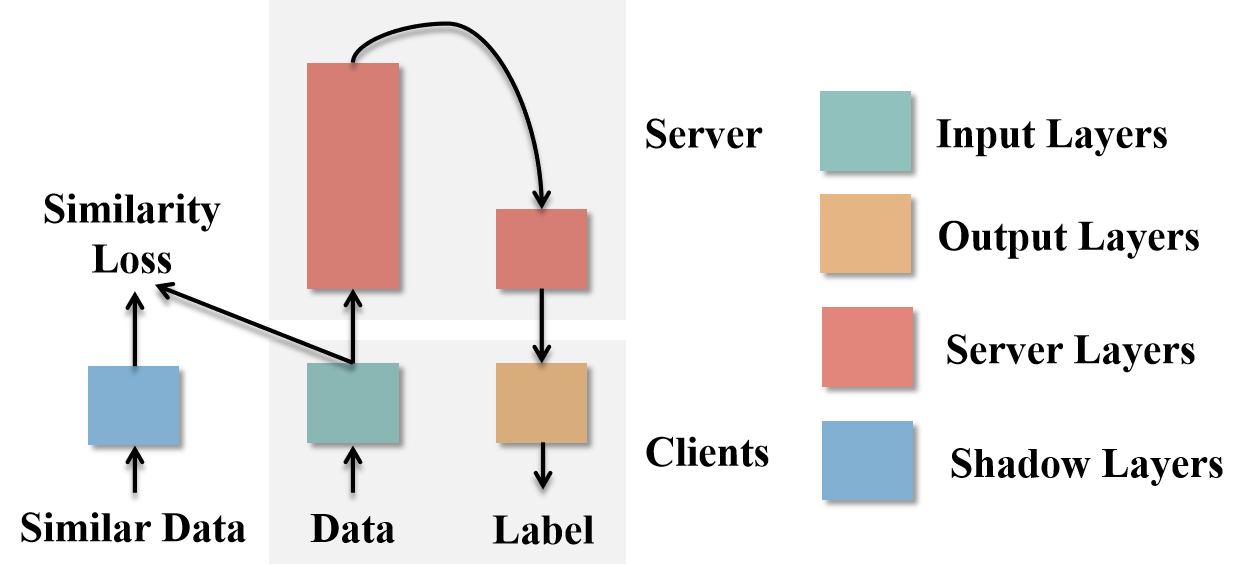}
    \caption{The sketch map of the shadow model training stage. We use shadow layers to substitute input layers. The resultant combination of shadow input layers and server layers serves as the proxy model. We leverage similarity loss to promote the similarity between input and shadow input layers for better attack effectiveness.}
    \label{shadow_model_training_fig}
\end{figure}

We elaborate on shadow model training and local adversarial attack of \sysname in Sections \ref{shadow_model_training} and \ref{local_attack}, respectively.
Algorithms \ref{shadow_model_training_algorithm} and \ref{local_adversarial_attack_algorithm} outline the two stages.
Figure \ref{shadow_model_training_fig} illustrates the overall idea of shadow model training stage.
Section \ref{theory} gives theoretical analysis on \sysname. Finally, 
Section \ref{detailed_comp} compares the difference between \sysname and common-used transfer-based attacks.

\subsection{Shadow Model Training}
\label{shadow_model_training}

\sysname trains a proxy model to launch attacks.
In split learning, the server holds server layers, which can be maliciously 
exploited by attackers to reduce the cost of launching attacks.
In response, a straightforward idea is to train shadow input layers and shadow output layers (compatible with server layers) and then concatenate them with server layers to form a proxy model.
However, attackers are unable to query ground-truth labels for the collected data, making it impossible to train the output layers without supervised signals.
Furthermore, even if the labels of the collected data are available, the label space of the collected data may differ from the label space of the target model.
This gap implies a low similarity between the proxy model and the target model, resulting in low attack success rates.
To overcome this issue, we attempt to discard the shadow output layers. As a result, crafting adversarial examples can only rely on the shadow input layers and server layers.
However, with the absence of an output layer, \sysname cannot generate adversarial examples by directly reducing the predicted probability of the corresponding label obtained from the proxy model. We defer the solution to this problem to Section \ref{local_attack}.

In split learning, clients send outputs of input layers to the server.
Intuitively, a simple idea is to adopt the outputs of input layers as supervised signals for shadow input layers.
In each iteration, an extra difference loss item, $L_2$-norm, is introduced, which measures the difference between the outputs of input layers and shadow input layers associated with their respective data.
In this way, shadow input layers can be trained.
It is stressed that this way does not alert clients.
Because the extra difference loss item is computed by the server, and, in each backpropagation process, the server just fuses the gradients of the original loss and gradients of the extra difference loss.
The original loss denotes the loss used to train models over the task, e.g., cross-entropy loss.
The fused gradients are returned to clients.
Clients cannot detect gradient tampering by only observing the returned gradients.
Formally, by denoting clients' and the server's data as $D_1$ and $D_2$, a forward process can be formally expressed as follows.

Clients transmit locally-computed output $o_1$ of input layers $F_{\theta_1}$ to the server: $o_1 = F_{\theta_1}(x), x \sim D_1$.
With $o_1$, the server returns the output $o_2$ of server layers $F_{\theta_2}$ associated with $o_1$ and compute the extra difference loss $L_{sim}$ to train shadow input layers:
\begin{equation}
\begin{split}
    o_2 &= F_{\theta_2}(o_1), \\ 
    L_{sim} &= ||F_{\theta_{1}'}(x')-o_1||_2, \ x' \sim D_2. \nonumber
\end{split}
\end{equation}
Afterwards, clients complete the rest of forward process, i.e., computing the final outputs of the model and the original loss:
\begin{equation}
    loss = L(F_{\theta_3}(o_2),y). \nonumber
\end{equation}

The back-propagation process is sequentially implemented in the following steps:
\begin{itemize}
    \item Clients update $\theta_3$ by $\frac{\partial loss}{\theta_3}$ and send $\frac{\partial loss}{o_3}$ back to the server;
    \item The server updates $\theta_2$ and $\theta_{1}'$ by $\frac{\partial loss}{\partial o_2} \frac{\partial o2}{\theta_2}$ and $\frac{L_{sim}}{\theta_{1}'}$ respectively, and sends $\frac{\partial loss}{\partial o_2} \frac{\partial o2}{\partial o_1} + \alpha \frac{\partial L_{sim}}{o_1}$ back to the clients, where $\alpha$ is similarity constraints;
    \item Clients update $\theta_1$ by $(\frac{\partial loss}{\partial o_2} \frac{\partial o2}{\partial o_1} + \alpha \frac{\partial L_{sim}}{o_1})\frac{\partial o_1}{\theta_1}$.
\end{itemize}

\par
Besides exploiting the difference between the outputs of input layers and shadow input layers as supervised signals, an alternative is to utilize the difference of outputs of server layers with respect to clients' data and the attacker's data.
However, the latter usually performs worse than the former.
In particular, it is well-known that the extracted features of the model transmit from general to specific along with the model depth \cite{adv_is_feature,deep_learning_survey}.
In light of the observation, due to the non-negligible data distribution divergence between clients and the server, forcefully aligning outputs of server layers probably causes collapse in model performance.
In contrast, it makes more sense to align the output of input layers since input layers typically extract features shared across different kinds of data.
We also validate this point in experiments (Section \ref{model_impact}).


\begin{algorithm}[t]
  \caption{Local Adversarial Attack}
  \label{local_adversarial_attack_algorithm}
  \begin{algorithmic}[1]
    \Require $F_{\theta_1'}$: shadow input layers;
             $F_{\theta_2}$: server layers.
             $x$: samples;
             $K$: the total number of iterations;
             $\epsilon$: perturbation budget.
      
    \Ensure $x+\delta$: the crafted adversarial examples.
    
    \State Initialize the adversarial noises $\delta$ for $x$.
     
    \For{each iteration $i = 1$ to $K$}
    
        \State Compute $o_2=F_{\theta_2}(F_{\theta_1'}(x))$ and $o_2'=F_{\theta_2}(F_{\theta_1'}(x+\delta))$.
        
        \State Compute loss $L_{attack}(o_2,o_2')=\frac{o_2 \cdot o_2'}{||o_2||_2 ||o_2'||_2}$.
    
        \State Optimize $\delta$ based on $L(o_2, o_2')$.

        \State Clip $\delta$ based on perturbation budget $\epsilon$.
        
    \EndFor
  \end{algorithmic}
\end{algorithm}

\subsection{Local Adversarial Attack}
\label{local_attack}

After shadow model training, the trained shadow input layers serve as a substitute for input layers.
We combine shadow input layers and server layers as the proxy model.
The problem here is that the proxy model lacks output layers, indicating that common techniques for crafting adversarial examples in transfer-based attacks cannot be directly implanted into our case as these methods require prediction probability. 
Therefore, instead of decreasing the prediction probability, \sysname crafts adversarial examples by perturbing the intermediate outputs.
Intuitively, since output layers make predictions based on the outputs of the server layers, it is possible to trick the target model by changing the outputs of the server layers.
Furthermore, if shadow input layers and input layers are similar enough, such noises that induce considerable shifts in outputs of the combination of the shadow input layers and server layers also are likely to produce consistent influences on outputs of the combination of the input layers and server layers.
Section \ref{theory} formally demonstrates the effectiveness of the above idea under some mild regularization conditions.

In practice, \sysname implements this idea by optimizing noises that reduce the cosine similarity between the intermediate outputs of perturbed and unperturbed samples.
Mathematically, by initializing $\delta$ as zero vectors, \sysname iteratively optimizes $\delta$ for $K$ times based on following steps:
\begin{itemize}
    \item Compute the output of server layers with respect to original and corresponding adversarial samples: $o_2 = F_{\theta_2}(F_{\theta_1'}(x)), o_2' = F_{\theta_2}(F_{\theta_1'}(x+\delta))$.
    \item Compute loss function that measures the similarity between $o_2$ and $o_2'$: $L_{attack}(o_2, o_2') =  \frac{o_2 \cdot o_2'}{||o_2||_2 ||o_2'||_2}$.
    \item Update $\delta$ by $\delta = Clip_{\epsilon}\{\delta - \beta \frac{\partial L_{attack}(o_2, o_2')}{\partial \delta}\}$, where $\beta$ is update step size, $\epsilon$ is given perturbation budget, and $Clip_{\epsilon}\{\cdot\}$ draws input into $\epsilon$-ball (usually $\infty$-norm) if input beyond perturbation budget.
\end{itemize}

\subsection{Detailed Comparison between \sysname and Commonly-used Transfer-based Attacks}
\label{detailed_comp}

Unlike commonly-used transfer-based attacks that focus on centralized training settings \cite{adv_survey,momen_attack}, \sysname instead pays more attention to decentralized settings, i.e., split learning, which is more suitable
in resource-restricted settings \cite{split_learning_1}.

In the proxy model training stage, aside from more loose needs for datasets, \sysname enjoys two extra advantages over commonly-used transfer-based attacks.
i) Attackers are assumed to own less information about the task performed by the target model, which is more practical as shown in Section \ref{attacker_knowledge}.
ii) \sysname is an immediate attack since the proxy and target models are trained together, suggesting that \sysname can launch attacks whenever the target model is deployed.

In the way of crafting adversarial examples, \sysname promotes the output of server layers w.r.t. crafted adversarial examples to have a large shift, rather than decreasing logits.
Moreover, experimental results (Section \ref{exp}) show that such method still can achieve competitive attack performance.

\section{Theoretical Analysis}
\label{theory}

Section \ref{evidence1} suggests that local adversarial attack is effective when the input layers and shadow layers are similar. 
Additionally, Section \ref{evidence2} demonstrates that adding an extra loss term (i.e., $L_2$-norm) can increase the similarity between the input layers and shadow layers. 

\subsection{The Effectiveness of Shadow Model Training}
\label{evidence1}

In the following analysis, we set $||\delta|| \leq \epsilon$ where $\epsilon$ is a small positive constant to make Taylor expansion established.
By adding $\delta$ into $x$, for the combination of shadow input layers and server layers, the output can be approximately estimated as follows:
\begin{equation}
\label{shadow_eq}
\begin{split}
    & F_{\theta_1'}(x+\delta) = F_{\theta_1'}(x) + \nabla_{x} F_{\theta_1'}(x)^T \cdot \delta = o_1' + \nabla_{x} F_{\theta_1'}(x)^T \cdot \delta,
    \\
    &F_{\theta_2}(F_{\theta_1'}(x+\delta))
    = F_{\theta_2}(o_1')+ \nabla_{o_1'} F_{\theta_2} (o_1')^T \cdot F_{\theta_1'}(x)^T \cdot \delta,
\end{split}
\end{equation}
where $o_1'$ denotes the output of the shadow input layers.
Similarly, for the combination of input layers and server layers, there is:
\begin{equation}
\label{normal_eq}
    F_{\theta_2}(F_{{\theta}_1}(x+\delta)) = F_{\theta_2}({o}_1)+ \nabla_{{o}_1} F_{\theta_2} (o_1)^T \cdot F_{{\theta}_1}(x)^T \cdot \delta.
\end{equation}

\sysname minimizes the cosine distance to craft $\delta$ as follows:
\begin{equation}
\label{optim_task}
    \begin{split}
        \delta &= \arg \min_{\delta} F_{\theta_2}(F_{\theta_1'}(x)) \cdot F_{\theta_2}(F_{\theta_1'}(x+\delta)) \\
        &= \arg \min_{\delta} F_{\theta_2}(o_1') \cdot ( F_{\theta_2}(o_1')+ \nabla_{o_1'} F_{\theta_2} (o_1')^T \cdot F_{\theta_1'}(x)^T \cdot \delta), \\
        &s.t., \quad ||\delta|| \leq \epsilon, ||F_{\theta_2}(F_{\theta_1'}(x))||=||F_{\theta_2}(F_{\theta_1'}(x+\delta))||=1.
    \end{split}
\end{equation}

The best solution of $\delta$ for the above optimization task is:
\begin{equation}
\begin{split}
        \delta &= - C \cdot F_{\theta_1'}(x) \cdot \nabla_{o_1'} F_{\theta_2} (o_1'), \\
        C &= \frac{\epsilon}{||F_{\theta_1'}(x) \cdot \nabla_{o_1'} F_{\theta_2} (o_1')||}.
\end{split}
\end{equation}

Then, by substituting the best solution of $\delta$ into Equation \ref{normal_eq}, it is observed that the influence of $\delta$ for input layers and server layers:
\begin{equation}
\begin{split}
        &F_{\theta_2}(F_{{\theta}_1}(x+\delta)) = F_{\theta_2}({o}_1)+ \nabla_{{o}_1} F_{\theta_2} (o_1)^T \cdot F_{{\theta}_1}(x)^T \cdot \delta
        \\&=  F_{\theta_2}({o}_1) -C \cdot \nabla_{{o}_1} F_{\theta_2} (o_1)^T \cdot F_{{\theta}_1}(x)^T \cdot F_{\theta_1'}(x) \cdot \nabla_{o_1'} F_{\theta_2} (o_1').
\end{split}
\end{equation}

The finding here is that if the inner dot between $F_{{\theta}_1}(x) \cdot \nabla_{{o}_1} F_{\theta_2} (o_1)$ and $F_{\theta_1'}(x) \cdot \nabla_{o_1'} F_{\theta_2} (o_1')$ is positive, $\delta$ can produce similar effects.
Specifically, when directions of the two items are totally same, $\delta$ induces identical effects.
However, there is no evidence to show the positive of the inner dot.
To solve this problem, $L_{sim}$ is added to increase the similarity between shadow input layers and input layers and we here demonstrate that $L_{sim}$ can promote direction consistency between the two gradient items.
\par
We first demonstrate $\nabla_{x} F_{\theta_1}(x)= 
\nabla_{x} F_{\theta_1'}(x), \forall x$ if $ F_{\theta_1}(x)=F_{\theta_1'}(x)$.
According to Lagrange mean value theorem, there is:
\begin{equation}
\begin{split}
    (F_{\theta_1}(x_1) - F_{\theta_1}(x_2)) (x-y)^{-1} &= \nabla_{\xi_1}F_{\theta_1}(\xi_1), x1 \leq \xi_1 \leq x2,\\
    (F_{\theta_1'}(x_1) - F_{\theta_1'}(x_2)) (x-y)^{-1} &= \nabla_{\xi_2}F_{\theta_1'}(\xi_2), x1 \leq \xi_2 \leq x2.
\end{split}
\end{equation}
Due to $ F_{\theta_1}(x)=F_{\theta_1'}(x), \forall x$, we have:
\begin{equation}
\begin{split}
    &(\nabla_{x_1} F_{\theta_1}(x_1) - \nabla_{x_2} F_{\theta_1}(x_2)) (x-y)^{-1} \\
    &= (\nabla_{x_1} F_{\theta_1'}(x_1) - \nabla_{x_2} F_{\theta_1'}(x_2)) (x-y)^{-1},\\
    &\nabla_{\xi_1}F_{\theta_1}(\xi_1) = \nabla_{\xi_2}F_{\theta_1'}(\xi_2).
\end{split}
\end{equation}
Let $x_2 \rightarrow x_1$, there is:
\begin{equation}
    x_1=\xi_1=\xi_2, \nabla_{x_1}F(\theta_1)(x_1) = \nabla_{x_1}F(\hat{\theta}_1)(x_1).
\end{equation}
Furthermore, $o_1=o_1'$ because of $ F_{\theta_1}(x)=F_{\theta_1'}(x)$.
Therefore, if shadow input layers and input layers output identically for all inputs, the directions of the two items aforementioned are same, which inspire us to add $L_{sim}$.
\par
Now, we relieve the condition $ F_{\theta_1}(x)=F_{\theta_1'}(x), \forall x$.
Let the maximum distance between $F_{\theta_1}(\cdot)$ and $F_{\theta_1'}(\cdot)$ equals to $d$ over its input space, i.e. $d=\max_{x} ||F_{\theta_1}(x)-F_{\theta_1'}(x)||, \forall x$.
Then, there is:
\begin{equation}
\begin{split}
&||\nabla F_{\theta_1}(x)- \nabla F_{\theta_1'}(x)|| \\
& \approx ||(F_{\theta_1}(x+\delta)-F_{\theta_1}(x)) \delta^{-1}  - (F_{\theta_1'}(x+\delta)-F_{\theta_1'}(x)) \delta^{-1}|| \\
&=  ||(F_{\theta_1}(x+\delta)-F_{\theta_1}(x) - F_{\theta_1'}(x+\delta)+ F_{\theta_1'}(x)) \delta^{-1}|| \\
&\leq \frac{1}{||\delta||} (||F_{\theta_1}(x+\delta) - F_{\theta_1'}(x+\delta)||+ ||F_{\theta_1}(x)- F_{\theta_1'}(x)||)\\
&= \frac{2d}{||\delta||}.
\end{split}
\end{equation}
As such, decreasing the distance of outputs of shadow input layers and input layers for same inputs still can enhance their gradient similarity.

\subsection{The Effectiveness of Local Adversarial Attack}
\label{evidence2}

The overall goal of this part is to study whether or not the samples perturbed by adding noises $\delta$ can fool the target model.
The target model is a combination of input layers, server layers, and output layers, and we discuss the impact of noises crafted by shadow input layers and server layers for the outputs of input layers and server layers before.
Therefore, the overall goal can be converted into the impact of the intermediate output changes $-C \cdot \nabla_{{o}_1} F_{\theta_2} (o_1)^T \cdot F_{{\theta}_1}(x)^T \cdot F_{\theta_1'}(x) \cdot \nabla_{o_1'} F_{\theta_2} (o_1')$ to output layers.
For the simplicity of symbols, $-C \cdot \nabla_{{o}_1} F_{\theta_2} (o_1)^T \cdot F_{{\theta}_1}(x)^T \cdot F_{\theta_1'}(x) \cdot \nabla_{o_1'} F_{\theta_2} (o_1')$ is denoted by $\Delta$.
We analyze the influence of $\Delta$ to loss function $L(\cdot)$ (the larger the loss function, the worse the target model performance):
\begin{equation}
    L(F_{\theta_3}(o_2+\Delta),y) = L(F_{\theta_3}(o_2),y) + \nabla_{o_2} L(F_{\theta_3}(o_2),y)^T \Delta.
\end{equation}
To maximize the above loss function, the directions of $\Delta$ should be aligned with $\nabla_{o_2} L(F_{\theta_3}(o_2),y)$ as possible.
\par
As shown in \cite{wiper,soteria}, the final part of DNNs usually exhibits high linearity, especially for the last fully-connected layer. 
Based on this, we can make the following approximation:
\begin{equation}
L(F_{\theta_3}(0),y)=L(F_{\theta_3}(o_2),y) + \nabla_{o_2}L(F_{\theta_3}(o_2),y) (0-o_2).
\end{equation}
Notice that, output layers for full-zero inputs also output zeros and then the loss $L(F_{\theta_3}(0),y)$ is huge.
Besides, the target model is generally trained well, implying $L(F_{\theta_3}(o_2),y)$ being small.
Thus, there is $\nabla_{o_2}L(F_{\theta_3}(o_2),y)^T o_2 < 0$.
\par
Furthermore, the best solution for $\delta$ is to minimize the cosine direction, thereby the inner product between $\Delta$ and $\hat{o}_2=F_{\theta_2}(F_{\theta_1}(x))$ being negative.
Moreover, $o_2$ and $\hat{o}_2$ probably are close with the aid of $L_{sim}$ shown in Part 1 and we assume $o_2^T \cdot \hat{o}_2 > 0$.
Therefore, there is $o_2^T \Delta < 0$ and $\nabla_{o_2}L(F_{\theta_3}(o_2),y)^T \Delta > 0$.
In short, if shadow input layers and input layers are similar, the noises crafted by \sysname can decrease the performance of the target model.



\section{Experimental Evaluation}
\label{exp}


\subsection{Setup}
According to existing works \cite{adv_survey}, three factors (i.e., the difference in architecture, training data distribution, and the numbers of training instances between the proxy and target models) are empirically demonstrated as being significant to attack effectiveness, while the leaving ones are usually trivial.
We primarily explore the robustness of split learning by varying these important factors (see Section \ref{data_impact_sec} and Section \ref{model_impact}) while keeping the trivial ones fixed at commonly-used values throughout evaluations~\cite{split_attack_1}.

\begin{table}[t!]
\caption{Accuracy drop (\%) of the target model trained over CIFAR-10 with 40,000 instances. The proxy model is trained over four different data distributions and varying numbers of data instances.}
\label{data_impact}
\centering
\small
\begin{tabular}{@{}c|cccc@{}}
\toprule
\# Instances & SVHN  & CIFAR-10 & CIFAR-100 & TinyImageNet \\ \midrule
128         & 14.07 & 16.31    & 20.31     & 15.47        \\
256         & 27.86 & 29.43    & 30.31     & 25.16        \\
1024        & 29.19 & 31.42    & 30.99     & 27.38        \\
2048        & 29.60 & 31.65    & 33.21     & 28.31        \\
4096        & 27.84 & 29.86    & 35.22     & 34.77        \\
8192        & 28.28 & 30.20    & 34.29     & 33.15        \\ \bottomrule
\end{tabular}
\end{table}

\begin{table}[t!]
\centering
\caption{Attack performance
of three attacks 
against the target model over CIFAR-10 measured by accuracy drop (\%).}
\label{comp_transfer}
\small
\begin{tabular}{@{}cccc@{}}
\toprule
\# Instances & VR & SSA & \sysname \\ \midrule
1024 & 19.15 & 20.88  & \textbf{31.42} \\
2048 & 22.48 & 24.60 & \textbf{31.65} \\
4096 & 23.71 & 25.43 & \textbf{29.86} \\ \bottomrule
\end{tabular}
\end{table}

\begin{table}[!t]
\caption{Impact of parameter $\alpha$.
With a fixed dataset size of 4096, we use different $\alpha$ to train proxy models in four different data distributions to implement \sysname against the target model.}
\label{sim_impact_asr}
\centering
\small
\begin{tabular}{@{}c|cccc@{}}
\toprule
$\alpha$  & SVHN & CIFAR-10 & CIFAR-100 & TinyImageNet \\ \midrule
0    & 10.87 & 12.58 & 15.15 & 14.79 \\
0.01 & 15.02 & 16.77 & 20.53 & 20.46 \\
0.1  & 21.58 & 21.33 & 26.45 & 25.04 \\
1    & 27.84 & 29.86 & 35.22 & 34.77 \\
10   & 28.28 & 30.52 & 38.01 & 35.88 \\
100  & 29.51 & 31.08 & 41.58 & 37.11 \\ \bottomrule
\end{tabular}
\end{table}

\begin{table}[t]
\caption{The impact of similarity constraints $\alpha$ of \sysname in terms of the accuracy of the target model.}
\label{sim_impact_acc}
\centering
\small
\begin{tabular}{@{}c|cccc@{}}
\toprule
\begin{tabular}[c]{@{}c@{}} $\alpha$ \end{tabular} & SVHN & CIFAR-10 & CIFAR-100 & TinyImageNet \\ \midrule
0    & 81.47 & 81.25 & 80.36 & 80.47 \\
0.01 & 81.81 & 80.58 & 82.03 & 80.47 \\
0.1  & 78.91 & 80.25 & 79.35 & 80.25 \\
1    & 78.45 & 78.12 & 78.23 & 78.01 \\
10   & 76.00 & 76.34 & 76.23 & 76.79 \\
100  & 75.33 & 75.67 & 77.34 & 75.89 \\ \bottomrule
\end{tabular}
\end{table}

\begin{table*}[!th]
\caption{The attack effectiveness of \sysname with different sizes of the shadow input layers and input layers.}
\label{model_depth_impact_table}
\centering
\small
\begin{tabular}{@{}c|ccc|ccc|ccc|ccc@{}}
\toprule
\multirow{2}{*}{Layer Depth} & \multicolumn{3}{c|}{SVHN} & \multicolumn{3}{c|}{CIFAR-10} & \multicolumn{3}{c|}{CIFAR-100} & \multicolumn{3}{c}{TinyImageNet} \\ \cmidrule(l){2-13} 
  & Single & Multi & Diff & Single & Multi & Diff & Single & Multi & Diff & Single & Multi & Diff \\ \midrule
1 & 27.84  & 27.84 & 0.00 & 29.86  & 29.86 & 0.00 & 35.22  & 35.22 & 0.00 & 34.77  & 34.77 & 0.00 \\
2 & 23.68  & 23.94 & 0.26 & 25.74  & 26.19 & 0.45 & 34.98  & 35.10 & 0.12 & 32.77  & 32.99 & 0.21 \\
3 & 18.55  & 20.48 & 1.93 & 21.79  & 22.90 & 1.11 & 32.11  & 33.76 & 1.66 & 28.56  & 30.20 & 1.64 \\
4 & 13.59  & 17.68 & 4.09 & 15.89  & 18.56 & 2.67 & 27.47  & 30.30 & 2.83 & 24.35  & 27.62 & 3.28 \\ \bottomrule
\end{tabular}
\end{table*}

\noindent
\textbf{Collaborative settings.}
A typical setting is considered here, where 10 clients collaboratively train a shared model under the coordination of the server.
Four state-of-the-art model architectures are selected for evaluation, i.e., MobileNet, DenseNet, ResNet18, and EfficientNet. The models are sequentially split into three parts to be allocated to clients and the server.
Moreover, the ResNet18 architecture serves as the default architecture of the shared model and we discuss the impact of architecture in Section \ref{model_impact}.
ResNet18 composes of 17 convolutional layers tailed with a fully-connected layer and input layers, server layers, and output layers hold 2, 15, and 1 layers.
Moreover, similar to \cite{split_attack_1}, we suppose that clients solve a benchmark image classification task of CIFAR-10 to evaluate and the attacker's data is randomly extracted from SVHN, CIFAR-10, CIFAR-100, and TinyImageNet datasets.
80\% of the training data in CIFAR-10 are evenly distributed to clients for training with a momentum optimizer with learning rate of 0.01 for 3000 iterations. 
In this stage, the server trains shadow input layers with similarity constraint $\alpha$ of 1 and an SGD optimizer with learning rate of 0.01.
Since the server (Section \ref{attacker_knowledge}) is assumed to have some knowledge about the AI domain and can infer the rough type of the current task, the default architecture of the shadow input model is set to two plain convolutional layers with skip connections.

\noindent
\textbf{Attack settings.}
In the attack stage, the server exploits the concatenation of the trained shadow input model and server layers as the proxy model.
We implement Algorithm \ref{local_adversarial_attack_algorithm} with perturbation budget of 0.3, step size of 0.3, and $K$ of 1 (small iterations can make cheaper attacks) to craft adversarial examples against the target model.

\noindent
\textbf{Metric.}
We assess the effectiveness of \sysname using accuracy drop, which is defined as the difference between the accuracy of the model in the presence of natural samples and adversarial samples, over the test set of CIFAR-10.
Higher accuracy drop indicates better attack performance.

\begin{table}[!th]
\caption{The impact of the target model architectures.
The proxy model is trained over four datasets with 4096 instances).
}
\label{model_arch_impact_table}
\centering
\small
\begin{tabular}{@{}c|cccc@{}}
\toprule
             & ResNet & EfficientNet & DenseNet & MobileNet \\ \midrule
SVHN         & 27.84  & 18.80        & 30.63    & 32.97     \\
CIFAR-10     & 29.86  & 28.28        & 39.56    & 38.45     \\
CIFAR-100    & 35.22  & 30.75        & 44.37    & 45.83     \\
TinyImageNet & 34.77  & 31.64        & 39.23    & 38.90     \\
Avg.         & 31.92  & 27.37        & 38.45    & 39.04     \\ \bottomrule
\end{tabular}
\end{table}

\subsection{The Impact of Data}
\label{data_impact_sec}
The training data of the proxy model is of crucial importance for attack effectiveness.
Generally, attack effectiveness grows with an increase in the number of available data and a decrease in the divergence between the data distributions of the target and proxy models.
This is because attack effectiveness heavily depends on the similarity of learned features between proxy and target models~\cite{adv_is_feature}.
A small volume of data makes it difficult to identify discriminative features while significant divergence in data distribution implies the underlying discriminative features behind these data are distinct.
Table \ref{data_impact} reports the results by altering the data number and data distribution of the attacker.
For SVHN, CIFAR-100, and TinyImageNet, we randomly extract data from the training dataset as the attacker's data. For CIFAR-10, we randomly sample the remaining training dataset (different from the data used for training the global model).


\noindent
\textbf{Data volume.}
The attack effectiveness considerably raises at the start of increasing data volume until it peaks at around 2000 to 4000, after which attack effectiveness tends to fluctuate moderately.
Surprisingly, we find that even with a tiny amount of data, \sysname can still achieve a non-trivial performance. 
Taking CIFAR-100 as an example, \sysname produces a 20\% accuracy drop in the target model by using 128 Non-IID data instances, highlighting the extreme vulnerability of split learning against adversarial attacks.

\noindent
\textbf{Data distribution.}
Compared to data volume, the impact of data distribution indicates more striking observations.
To begin with, intuitively, the best attack effectiveness should be achieved when the attacker's data is similar to the training data of the target model, i.e., CIFAR-10.
However, Table \ref{data_impact} shows the attack effectiveness of employing different datasets as training data for the shadow input layers as follows: CIFAR100 $>$ CIFAR10 $>$ TinyImageNet $>>$ SVHN.
This phenomenon can be explained by the fact that early layers primarily capture low-level features \cite{adv_is_feature}, which are shared among these datasets despite the divergence in their high-level features.
Notably, the attack effectiveness using SVHN is slightly lower than that of the other datasets.
We speculate this is because SVHN is a digital image dataset, and consequently, its low-level feature distribution differs from animal or plant images (CIFAR-10, CIFAR-100, and TinyImageNet). In summary, the quality requirement of the training data distribution for shadow input layers to launch adversarial attacks is lower than we thought.


\subsection{Comparison to State-of-the-art Attacks}

We compare the attack performance of \sysname with two state-of-the-art transfer-based attacks, namely VR~\cite{trans_adv_VR} and SSA~\cite{trans_adv_SSA}.
For VR and SSA, we train a proxy model of the same architecture of the target model in CIFAR-10 with various sizes of training sets (the same training dataset of \sysname) and then craft adversarial examples using the hyperparameters suggested in their original paper.
Table \ref{comp_transfer} reports the attack effectiveness of the three attacks.
Wherein, the attack performance of \sysname surpasses both baselines by a large margin, regardless of the dataset size.
In addition, the proxy model training of VR and SSA uses labels.
Therefore, \sysname is a more powerful and cost-effective attack than baselines.

\subsection{The Impact of Similarity Constraints}
\label{sim_impact}

Section \ref{theory} presents theoretical evidence that adding similarity constraints 
$\alpha$ for input and shadow input layers can improve attack effectiveness, which is further verified in this section.
Table \ref{sim_impact_asr} reports the attack effectiveness of \sysname over varying magnitudes of similarity constraints.
As shown in Table \ref{sim_impact_asr}, increasing the magnitude of similarity constraints substantially strengthens the attack effectiveness, demonstrating consistency with our theoretical analysis.

However, we emphasize that increasing the magnitude of similarity constraints also has a negative side.
Specifically, Table \ref{sim_impact_acc} shows the accuracy of the shared model over the test set at different values of similarity constraints.
As can be seen, larger similarity constraints lead to greater loss in model performance.



\subsection{The Impact of Model}
\label{model_impact}


\noindent \textbf{The impact of input layer depth.}
By holding shadow input layers fixed while incrementally increasing input layer depth, we study the impact of the depth of the input layers on the attack effectiveness of \sysname.
Table \ref{model_depth_impact_table} reports the attack effectiveness for the shared model with input layers of different depths, where we use Single to denote the fixed shadow input layers.
Overall, an increase in the depth of input layers leads to a decrease in the attack effectiveness of \sysname and this can be attributed to the model similarity.
On the one hand, increasing the depth of input layers makes a bigger difference between input and shadow input layers.
On the other hand, input layers with big depth tend to capture high-level features rather than low-level features~\cite{survey_cnn}.
To validate this point, we increase shadow input layers denoted in Multi in Table \ref{model_depth_impact_table}.
Using shadow input layers with a deeper depth can partly offset the effect of increasing the depth of input layers but cannot completely offset.

\noindent \textbf{The impact of target model architecture.}
We change the architecture of the target model and then implement \sysname to further examine the effectiveness of \sysname.
Table \ref{model_arch_impact_table} reports the attack performance of \sysname against the target model of different architectures.
Simply speaking, \sysname can effectively attack four model architectures.
Moreover, the results in Table \ref{model_arch_impact_table} indicate that DenseNet and MobileNet exhibit a higher vulnerability to adversarial attacks when compared to ResNet and EfficientNet.
This observation aligns with findings reported in baselines (VR and SSA).
A conjecture is that the simpler and more lightweight design of MobileNet renders it less robust against attacks (simple architectures may be not able to learn enough discriminative features).
On the other hand, DenseNet's distinctive characteristic of having more skip connections allows errors from earlier layers to propagate more easily to the last layer, potentially increasing its vulnerability to attacks.









\section{Conclusion}
\label{conclusion}
We designed an attack method called \sysname to sufficiently harness the characteristics of split learning.
In detail, \sysname comprises two stages: shadow model training and local adversarial attack. Shadow model training enables attackers to train shadow input layers inexpensively, compensating for the lack of input layers. Local adversarial attack employs shadow input layers and server layers to generate adversarial examples against the target model (i.e., the shared model). Our study also includes a thorough theoretical analysis of \sysname's attack effectiveness.
Extensive empirical experiments demonstrated the superior attack performance of \sysname and reveal the striking vulnerability of split learning to adversarial attacks.


\clearpage

\ack This work was supported by the National Natural Science Foundation of China under grant number 62202170 and Alibaba Group through the Alibaba Innovation Research Program.

\bibliography{ecai}

\end{document}